\documentclass[11pt,a4paper]{article}
\usepackage[hyperref]{emnlp2018}
\usepackage{times}
\usepackage{latexsym}
\usepackage{arydshln}
\usepackage{url}
\usepackage{xcolor}
\usepackage{comment}
\usepackage{graphicx}
\usepackage{mathtools}
\usepackage{amsmath}

\aclfinalcopy 
\def\aclpaperid{124} 

\newcommand\short[1]{{}}

\newcommand\replace[1]{{}}

\title{Neural Machine Translation into Language Varieties}
   
\author{ Surafel M. Lakew$^{\dagger \star}$, Aliia Erofeeva$^{\dagger}$, Marcello Federico$^{\star +}$   \\ $^{\dagger}$University of Trento, $^{\star}$Fondazione Bruno Kessler, $^{+}$MMT Srl, Trento, Italy \\
   {\tt $^{\dagger}$name.surname@unitn.it, $^{\star}$surname@fbk.eu} \\}

\date{}

\begin{document}
\maketitle
\begin{abstract}
Both research and commercial machine translation have so far neglected the importance of properly handling the spelling, lexical and grammar divergences occurring among language varieties. Notable cases are standard national varieties such as Brazilian and European Portuguese, and Canadian and European French, which popular online machine translation services are not keeping distinct. We show that an evident side effect of modeling such varieties as unique classes is the generation of inconsistent translations. In this work, we investigate the problem of training neural machine translation from English to specific pairs of language varieties, assuming both labeled and unlabeled parallel texts, and low-resource conditions. We report experiments from English to two pairs of dialects, European-Brazilian Portuguese and European-Canadian French, and two pairs of standardized varieties, Croatian-Serbian and Indonesian-Malay.\short{Each target pair representing different levels of linguistic similarity as well as different data and label distributions. We compare dialect/language specific and generic NMT baselines against multilingual NMT systems exploiting manual as well as automatic language labels.} We show significant BLEU score improvements over baseline systems when translation into similar languages is learned as a multilingual task with shared representations. 

\end{abstract}

\section{Introduction}
The field of machine translation (MT) is making amazing progress, thanks to the advent of neural models and deep learning. While just few years ago research in MT was struggling to achieve {\em useful} translations for the most requested and high-resourced languages, the level of translation quality reached today has raised the demand and interest for less-resourced languages and the solution of more subtle and interesting translation tasks~\cite{bentivogli:CSL2018}.  If the goal of machine translation is to help worldwide communication, then the time has come to also cope with dialects or more generally language varieties\footnote{In sociolinguistics, a variety is a specific form of language, that may include dialects, registers, styles, and other forms of language, as well as a standard language. See \newcite{Wardhaugh} for a more comprehensive introduction.}. Remarkably, up to now, even standard national language varieties, such as Brazilian and European Portuguese, or Canadian and European French, which are used by relatively large populations have been quite neglected both by research and industry. Prominent online commercial MT services, such as Google Translate and Bing, are currently not offering any variety of Portuguese and French. Even worse, systems offering such languages tend to produce inconsistent outputs, like mixing lexical items from different Portuguese (see for instance the translations shown in  Table~\ref{table:examples}).  Clearly, in the perspective of delivering high-quality MT to professional post-editors and final users, this problem urges to be fixed.

While machine translation from many to one varieties is intuitively simpler to approach\footnote{We will focus on this problem in future work and disregard  possible varieties in the source side, such as American and British English, in this work.}, it is the opposite direction that presents the most relevant problems. First,  languages varieties such as dialects might significantly overlap thus making differences among their texts quite subtle (e.g., particular grammatical constructs or lexical divergences like the ones reported in the example). Second, parallel data are not always labeled at the level of language variety, making it hard to develop specific NMT engines. Finally, training data might be very unbalanced among different varieties, due to the population sizes of their respective speakers or for other reasons. This clearly makes it harder to model the lower-resourced varieties~\cite{koehn2017six}.

\begin{table*}[t]
\centering
\small
\begin{tabular}{p {0.2 \textwidth} p{0.75 \textwidth}}
\hline
English (source)  & I'm going to the \underline{gym} before \underline{breakfast}. No, I'm not going to the \underline{gym}. \\[1mm]
\hline
pt (Google Translate) & Eu estou indo para a \textcolor{olive}{academia} antes do \textcolor{olive}{caf\'e da manh\~a}. N\~ao, eu n\~ao vou ao \textcolor{blue}{gin\'asio}.\\ 
pt-BR (M-C2)& Eu vou \'a \textcolor{olive}{academia} antes do \textcolor{olive}{caf\'e da manh\~a}.
N\~ao, eu n\~ao vou \`a \textcolor{olive}{academia}.\\ 
pt-EU (M-C2) & Vou para o \textcolor{blue}{gin\'asio} antes do \textcolor{blue}{pequeno-almo\c{c}o}.
N\~ao, n\~ao vou para o \textcolor{blue}{gin\`asio}. \\ 
pt-BR (M-C2\_{\tt L}) & Vou \`a \textcolor{olive}{academia} antes do \textcolor{olive}{caf\'e da manh\~a}. N\~ao, n\~ao vou \`a \textcolor{olive}{academia}. \\ 
pt-PT (M-C2\_{\tt L}) & Vou ao \textcolor{blue}{gin\'asio} antes do \textcolor{blue}{pequeno-almo\c{c}o}. N\~ao, n\~ao vou ao \textcolor{blue}{gin\'asio}.  \\ 
\hline

\end{tabular}
\caption{MT from English into Portuguese varieties. Example of mixed translations generated by Google Translate (as of 20th July, 2018) and translations generated by our variety-specific  models. For the underlined English terms both their \textcolor{olive}{Brazilian} and \textcolor{blue}{European}  translation variants are shown.}
\label{table:examples}
\end{table*}

In this work we present our initial effort to systematically investigate ways to approach NMT from English into four pairs of language varieties: Portuguese European - Portuguese Brazilian, European French - Canadian French, Serbian - Croatian, and Indonesian - Malay\footnote{According to Wikipedia, Brazilian Portuguese is a dialect of European Portuguese, Canadian French is a dialect of European French, Serbian and Croatian are standardized registers of Serbo-Croatian, and Indonesian is a standardized register of Malay.}. For each couple of  varieties, we assume to have both parallel text labeled with the corresponding couple member, and  parallel text without such information. 
Moreover, the considered target pairs, while all being mutually intelligible, present different levels of linguistic similarity and also different proportions of available training data. For our tasks we rely on the WIT$^3$ TED Talks collection\footnote{http://wit3.fbk.eu/},  used for the International Workshop of Spoken Language Translation, and OpenSubtitles2018, a corpus of subtitles available from the OPUS collection\footnote{http://opus.nlpl.eu/}.  

After presenting related work (Section~2) on NLP and MT of dialects and related languages, we introduce (in Section~3) baseline NMT systems, either language/dialect specific or generic, and multilingual NMT systems, either trained with fully supervised (or labeled) data or with partially supervised data. In Section~4, we introduce our datasets, NMT set-ups based on the Transformer architecture, and then present the results for each evaluated system. We conclude the paper with a discussion and conclusion in Sections~5 and 6.

\section{Related work}
\subsection{Machine Translation of Varieties}
Most of the works on translation between and from/to written language varieties involve rule-based transformations, e.g., for European and Brazilian Portuguese~\cite{Marujo2011}, Indonesian and Malay~\cite{Tan2012}, Turkish and Crimean Tatar~\cite{Altintas2002}; 
or phrase-based statistical MT (SMT) systems, e.g., for Croatian, Serbian, and Slovenian~\cite{Popovic2016}, Hindi and Urdu~\cite{durrani2010hindi}, or Arabic dialects~\cite{Harrat2017}. 
Notably, \newcite{Pourdamghani2017} build an unsupervised deciphering model to translate between closely related languages without parallel data. 
\newcite{Salloum2014} handle mixed Arabic dialect input in MT by using a sentence-level classifier to select the most suitable model from an ensemble of multiple SMT systems.
In NMT, however, there have been fewer studies addressing language varieties. 
It is reported that an RNN model outperforms SMT when translating from Catalan to Spanish~\cite{Costajussa2017} and from European to Brazilian Portuguese~\cite{Costa-Jussa2018}.
\newcite{Hassan2017} propose a technique to augment training data for under-resourced dialects via projecting word embeddings from a resource-rich related language, thus enabling training of dialect-specific NMT systems. The authors generate spoken Levantine-English data from larger Arabic-English corpora and report improvement in BLEU scores compared to a low-resourced NMT model.

\subsection{Dialect Identification}
A large body of research in dialect identification stems from the DSL shared tasks~\cite{Zampieri2014,Zampieri2015,Malmasi2016,Zampieri2017}. 
Currently, the best-performing methods include linear machine learning algorithms such as SVM, na\"ive Bayes, or logistic regression, which use character and word $n$-grams as features and are usually combined into ensembles~\cite{Jauhiainen2018}. 
\newcite{Tiedemann2012} present the idea of leveraging parallel corpora for language identification: content comparability allows capturing subtle linguistic differences between dialects while avoiding content-related biases. 
The problem of ambiguous sentences, i.e., those for which it is impossible to decide upon the dialect tag, has been demonstrated for Portuguese by \newcite{Goutte2016} through inspection of disagreement between human annotators.

\subsection{Multilingual NMT}
In a \textit{one-to-many} multilingual translation scenario, \newcite{dong2015multi} proposed a multi-task learning approach that utilizes a single encoder for source languages and separate attention mechanisms and decoders for every target language. 
\newcite{luong2015multi} used distinct encoder and decoder networks for modeling language pairs in a \textit{many-to-many} setting. 
\newcite{firat2016multi} introduced a way to share the attention mechanism across multiple languages. 
A simplified and efficient multilingual NMT approach is proposed by \newcite{johnson2016google} and \newcite{ha2016toward}  
by prepending language tokens to the input string. This approach has greatly simplified multi-lingual NMT, by eliminating the need of having separate encoder/decoder networks and attention mechanism for every new language pair.  In this work we follow a similar strategy, by incorporating an artificial token as a unique \textit{variety flag}.

\section{NMT into Language Varieties}\label{proposed_approach}
Our assumption is to translate from language $E$ (English) into each of two varieties $A$ and $B$. We assume to have parallel training data $D_{E\rightarrow A}$ and $D_{E\rightarrow B}$ for each variety as well as unlabeled data $D_{E\rightarrow A\cup B}$. For the sake of experimentation we consider three application scenarios in which a fixed amount of parallel training data $E$-$A$ and $E$-$B$ is partitioned in different ways: 

\begin{itemize}
\item {\em Supervised}: all sentence pairs are respectively put in $D_{E\rightarrow A}$ and $D_{E\rightarrow B}$, leaving $D_{E\rightarrow A\cup B}$ empty;
\item {\em Unsupervised}: all sentence pairs are jointly put in  $D_{E\rightarrow A\cup B}$, leaving $D_{E\rightarrow A}$ and $D_{E\rightarrow B}$ empty;
\item {\em Semi-supervised}: two-third of $E$-$A$ and $E$-$B$ are, respectively, put in $D_{E\rightarrow A}$ and $D_{E\rightarrow B}$, and the remaining sentence pairs are put in $D_{E\rightarrow A\cup B}$.
\end{itemize}

\short{In this work, we also assume to work with an encoder-decoder-attention architecture, in particular the Transformer model~\cite{vaswani2017attention}.} 

\noindent
{\bf Supervised and Unsupervised Baselines.} For each translation direction  we compare three baseline NMT systems. The first system is an unsupervised generic ({\tt Gen}) system trained on the union of the language varieties training data. Notice that {\tt Gen} makes no distinction between $A$ and $B$ and uses all data in an unsupervised way.  The second is a supervised variety-specific system ({\tt Spec}) trained on the corresponding language variety training set. The third system ({\tt Ada}) is obtained by adapting the {\tt Gen} system to a specific variety.
\footnote{We test this system only on the Portuguese varieties.} 
Adaptation is carried out by simply restarting the training process from the generic model using all the available variety specific training data. 

\noindent
{\bf Supervised Multilingual NMT.}
We build on the idea of multilingual NMT ({\tt Mul}), where one single NMT system is trained on the union of $D_{E\rightarrow A}$ and $D_{E\rightarrow B}$.  Each source sentence both at training and inference time is prepended with the corresponding target language variety label ($A$ or $B$). Notice that the multilingual architecture leverages the target forcing symbol both as input to the encoder to build its  states, and as initial input to the decoder to trigger the first target word. 

\noindent
{\bf Semi-Supervised Multilingual NMT.}
We consider here multilingual NMT models that make also use of unlabeled data  $D_{E\rightarrow A\cup B}$.
The first model we propose, named {\tt M-U}, uses the available data $D_{E\rightarrow A}$, $D_{E\rightarrow B}$ and $D_{E\rightarrow A\cup B}$ as they are, by not specifying any
label at training time for entries from $D_{E\rightarrow A\cup B}$.  The second model, named {\tt M-C2}, works
similarly to {\tt Mul},  but relying on a language variety identification module (trained on the target data 
of $D_{E\rightarrow A}$ and $D_{E\rightarrow B}$) that maps each unlabeled data point either to $A$ or $B$. The third model, named {\tt M-C3}, can be seen as an enhancement of {\tt M-U}, as the unlabeled data is automatically classified into one of three classes: $A$, $B$, or $A\cup B$. For the third class, like with {\tt M-U}, no label is applied on the source sentence.

\section{Experimental Set-up}
\subsection{Dataset and Preprocessing}
The experimental setting consists of eight target varieties and English as source. We use publicly available datasets from the WIT$^3$ TED corpus~\cite{cettoloEtAl:EAMT2012}. The summary of the partitioned training, dev, and test sets are given in Table~\ref{table:data}, where Tr. 2/3 is the labeled portion of the training set used to train the semi-supervised models, while the other 1/3 are either held out as unlabeled (\texttt{M-U}) or classified automatically (\texttt{M-C2}, \texttt{M-C3}).
In the preprocessing stages, we tokenize the corpora and remove lines longer than 70 tokens. The Serbian corpus written in Cyrillic is transliterated into Latin script with CyrTranslit\footnote{https://pypi.org/project/cyrtranslit}.\short{ For the mixed and multilingual models we concatenate the dataset according to the dialect categories.}
In addition, to also run a large-data experiment, we expand the 
English$-$European/Brazilian Portuguese
data with the corresponding\short{ publicly available}{} OpenSubtitles2018 datasets from the OPUS corpus. Table~\ref{table:data} summarizes the augmented training data, while keeping the same dev and test sets.

\begin{table}[t!]
\small
\centering
\begin{tabular}{rrrrrr}
\hline
		 &Train & Ratio (\%)  & Tr. 2/3 & Dev & Test  \\
  \hline
  pt-BR & 234K & 58.23 & 156K & 1567 & 1454 \\
  pt-EU & 168K & 47.77 & 56K & 1565 &  1124 \\ 
\hdashline
  fr-CA & 18K & 10.26 & 12K & 1608 & 1012\\
  fr-EU & 160K & 89.74 & 106K & 1567 & 1362\\ 
  \hdashline
  hr & 110K & 54.20 & 73K & 1745 & 1222\\
  sr & 93K & 45.80 &  62K & 1725 & 1214\\ 
 \hdashline
  id & 105k & 96.71& 70K & 932 &  1448 \\
  ms & 3.6K & 3.29 & 2.4k & 1024 & 738\\ 
  \hline
 pt-BR\_L & 47.2M & 64.91 & 31.4M & 1567 & 1454 \\
 pt-EU\_L & 25.5M & 35.10 & 17M & 1565 &  1124 \\ 
  \hline
\end{tabular}
\caption{Number of parallel sentences of the TED Talks used for training, development and testing. At the bottom, the large-data set-up which uses the OpenSubtitles (pt-BR\_L and pt-PT\_L) as additional training set.}
\label{table:data}
\end{table}

\subsection{Experimental Settings}
We trained all systems using the Transformer model\footnote{https://github.com/tensorflow/tensor2tensor}~\cite{tensor2tensor}.
We use the Adam optimizer~\cite{kingma2014adam} with an initial learning rate of $0.2$ and a dropout also set to $0.2$. 
A shared source and target vocabulary of size 16k is generated via sub-word segmentation~\cite{wu2016google}. The choice for the vocabulary size follows the recommendations in \newcite{denkowski2017stronger} regarding training of NMT systems on TED Talks data.  
Overall we use a uniform setting for all our models, with a $512$ embedding dimension and hidden units, and 6 layers of self-attention encoder-decoder network. The training batch size is of $6144$ sub-word tokens and the max length after segmentation is set to $70$.\short{ At inference time, we employ a beam size of $4$ and a batch size of $32$.} 
Following \newcite{vaswani2017attention} and for a fair comparison, experiments are run for 100k training steps, i.e., in the low-resource settings all models are observed to converge within these steps. 
Adaptation experiments are run to convergence, which requires roughly half of the steps (i.e., 50k) required to train the generic low-resource model.
On the other hand, large-data systems are trained for up to 800k steps, which also showed to be a convergence point. For the final evaluation we take the best performing checkpoint on the dev set. All models are trained using Tesla V100-pcie-16gb on a single GPU.

\subsection{Language Variety Identification}
To automatically identify the language variety of unlabeled target sentences, we train a fastText model~\cite{Joulin2017}, a simple yet efficient linear bag of words classifier. 
We use both word- and character-level $n$-grams as features.
In the low-resource condition, we train the classifier on the 2/3 portion of the labeled training data. For the large-data experiment, instead, we used a relatively smaller and independent corpus consisting of 3.3 million pt-BR$-$pt-EU parallel sentences extracted from OpenSubtitles2018 after filtering out identical sentences pairs and sentences occurring (in any of the two varieties) in the NMT training data.
Additionally, low-resource training sentences (fr-CA and ms) are randomly oversampled to mitigate class imbalance.

\begin{table}[t!]
\small
\centering
\begin{tabular}{llllll}
\hline
				& pt 			& sr-hr 			& fr 		& id-ms 	& pt\_{\tt L} \\
\hline
ROC AUC 		& 82.29 		& 88.12& 80.99 		& 81.99		& 52.75 		\\
\hline
\end{tabular}
\caption{Performance of language identification on the low-resource and high-resource (pt\_{\tt L}) settings}
\label{table:class-acc}
\end{table}

For each pair of varieties, we train five base classifiers differing in random initialization.
In the {\tt M-C2} experiments, prediction is determined based on soft fusion voting, i.e., the final label is the argmax of the sum of class probabilities.
Due to class skewness in the evaluation set, we report binary classification performance in terms of ROC AUC~\cite{Fawcett2006} instead of accuracy in Table~\ref{table:class-acc}.
For {\tt M-C3} models, we handle ambiguous examples using the majority voting scheme: in order for a label to be assigned, its softmax probability should be strictly higher than fifty percents according to the majority of the base classifiers, otherwise no tag is applied. On average, this resulted in $<$1\% of unlabeled sentences for the small data condition, and about 2\% of unlabeled sentences for the large data condition.

\section{Results and Discussion}
We run experiments with all the systems introduced in Section \replace{~3}{\ref{proposed_approach}}, on four pairs of languages varieties. Results are reported in Table~\ref{table:res} for the low-resource setting and in Table~\ref{table:res_large} for the large data setting. 

\subsection{Low-resource setting}
\begin{table}[t!]
\centering
\begin{tabular}{llrrr}
\hline
        &  & pt-BR & pt-EU & average  \\
  \hline
Unsuper.   &  Gen & $\downarrow$36.52 & $\downarrow$33.75 &  35.14\\ 
Supervis.	& Spec & $\downarrow$35.85  &  $\downarrow$35.84 & 35.85 \\
\hspace{0.5cm}"     & Ada &  $\downarrow$36.54	& $\downarrow$36.59	& 36.57	\\ 
\hspace{0.5cm}"		&  Mul & {\bf 37.86} & {\bf 38.42} & {\bf 38.14}\\
Semi-sup.   &  M-U & $\downarrow$37.09 & 37.59 & 37.34 \\ 
\hspace{0.5cm}"	&  M-C2 & {\bf 37.70} & {\bf 38.35} & {\bf 38.03} \\
\hspace{0.5cm}"  &  M-C3 & 37.59 & 38.31 & 37.95\\[2.5mm]
\hline
         &  & fr-EU & fr-CA & average  \\
  \hline
Unsuper.	    &  Gen & {\bf 33.91} & $\downarrow$30.91 & 32.41\\
Supervis.  & Spec & 33.52 & $\downarrow$17.13 & 25.33 \\ 	       
\hspace{0.5cm}" 	         &  Mul & 33.40  & {\bf 37.37} & {\bf 35.39} \\

Semi-sup.   & M-U & 33.28  & 37.96 & 35.62 \\
\hspace{0.5cm}"        	& M-C2 & 33.79 & $\uparrow$38.60 & 36.20 \\
\hspace{0.5cm}"   			& M-C3 & $\uparrow${\bf 34.16} & $\uparrow${\bf 39.30} & {\bf 36.73} \\[2.5mm]
 \hline
         &  & hr & sr & average  \\
  \hline
Unsuper. 		   &  Gen &  $\downarrow$21.71 & $\downarrow$19.20 & 20.46 \\
Supervis.   & Spec & $\downarrow$22.50 & $\downarrow$19.92 & 21.21 \\ 	        
\hspace{0.5cm}" 		     &  Mul & {\bf 23.99} & {\bf 21.37} & {\bf 22.68}\\

Semi-sup.   & M-U & {\bf 24.30} & 21.53 & 22.91 \\
\hspace{0.5cm}"			& M-C2 & 24.14 & 21.26 & 22.70\\
\hspace{0.5cm}"   			& M-C3 & 24.22 & {\bf 21.97} & {\bf 23.10}\\[2.5mm]
 \hline
         &  & id & ms & average  \\
  \hline
Unsuper.	   &  Gen & 26.56 & $\downarrow$13.86 & 20.21\\
Supervis.   & Spec & 26.20 & $\downarrow$2.73 & 14.47 \\
 	        
\hspace{0.5cm}"   			&  Mul &  {\bf 26.66} & {\bf 15.77} & {\bf 21.22}\\

Semi-sup.   & M-U & {\bf 26.52} & 15.58 &  21.05\\
\hspace{0.5cm}"			& M-C2 & 26.36 & {\bf 16.31} & {\bf 21.34}\\
\hspace{0.5cm}"   			& M-C3 & 26.40 & 15.23 & 20.82\\
 \hline
\end{tabular}
\caption{BLEU scores of the presented models, trained with unsupervised, supervised and semi-supervised data, from English to Brazilian Portuguese (pt-BR) and European Portuguese (pt-EU), Canadian French (fr-CA) and European French (fr-EU), Croatian (hr) and Serbian (sr), and Indonesian (id) and Malay (ms).
Arrows $\downarrow\uparrow$ indicate statistically significant differences calculated against \texttt{Mul} using bootstrap resampling with $\alpha=0.05$~\cite{Koehn2004}.
}
\label{table:res}
\end{table}

Among the supervised models, which are using all the available training data, the multilingual NMT model {\tt Mul} outperforms the variety-specific models on all considered directions. Remarkably, the   {\tt Mul} model also outperforms the adapted {\tt Ada} model on the available translation directions.  The unsupervised generic model {\tt Gen}, that mixes together all the available data, as expected tends to perform better 
than the supervised specific models of the less resourced varieties. Particularly, this improvement is observed for Malay (ms) and Canadian French (fr-CA), which respectively represent the 3.3\% and 10\% of the overall training data 
used by their corresponding ({\tt Gen}) systems. On the contrary, a  
degradation is observed for  European Portuguese (pt-Eu) and Serbian (sr), which 
represent 42\% and 45\% of their respective training sets. Even though very low-resourced varieties can benefit from the mix, it is also evident that the {\tt Gen} model can easily get biased because of the imbalance between the datasets.  

In the semi-supervised scenario, we report results with
three multilingual systems that integrate the 1/3 of unlabeled data to the training corpus in three different ways: {\em (i)}~without labels ({\tt M-U}), {\em (ii)}~with automatic labels forcing one of two possible classes ({\tt M-C2}), {\em (iii)}~with automatic labels of one of the two options or no label in case of low confidence of the classifier ({\tt M-C3}).

Results show that on average automatic tagging of the unlabeled data is better than leaving them unlabeled, although {\tt M-U} still remains a better choice than using specialized and generic systems. The best between {\tt M-C2} and {\tt M-C3} performs on average from very close to better than the best supervised method. 

If we look at the single language variety, the obtained figures are not showing a coherent picture. In particular, in the Croatian-Serbian and Indonesian-Malay pairs the best resourced language seems to benefit more from keeping the data unlabeled ({\tt M-U}). Interestingly, even the worst semi-supervised model performs very close or even better than the best supervised model, which suggests the importance of taking advantage of all available data even if they are not labeled. 

Focusing on the statistically significant improvements, the best supervised ({\tt Mul}) is better than the unsupervised ({\tt Gen}), whereas the best semi-supervised ({\tt M-C2} or {\tt M-C3}) is either comparable or better than the best supervised.

\subsection{High-resource setting}
Unlike what observed in the low-resource setting, where {\tt Mul} outperforms {\tt Spec} in the supervised scenario, in the large data condition, variety specific models apparently seem the best choice\short{ for the same evaluation set as in the low-resourced task}. Notice, however, that the supervised multilingual system {\tt Mul} provides just a slightly lower level of performance with a simpler architecture (one network in place of two). The unsupervised generic model {\tt Gen}, trained with the mix of the two varieties datasets, performs significantly worse than the other two supervised approaches, this is particularly visible for the pt-EU direction. Very likely, in addition to the ambiguities that arise from naively mixing the data of the two different dialects, there is  
also a bias effect towards pt-BR which is due to the very unbalanced  proportions of data between the two dialects (almost 1:2). 

Hence, in the considered high-resource setting,  the {\tt Spec} and {\tt Mul} models result as best possible solutions against which comparing our semi-supervised approaches.

\begin{table}[t!]
\centering
\begin{tabular}{llrrr}
\hline
				&				& pt-BR & pt-EU				& average		\\
\hline
Unsuper.		& Gen		& $\downarrow$ 39.78			& $\downarrow$ 36.13			& 37.96 		\\ \cdashline{2-5}
Supervis.		& Spec 		& {\bf 41.54} 	& {\bf 40.42} 	& {\bf 40.98} 	\\  		
\hspace{0.5cm}"	& Mul		& 41.28 		& 40.28			& 40.78			\\ \cdashline{2-5} 
Semi-sup.		& M-U		& 41.21 		& 39.88			& 40.55			\\ 
\hspace{0.5cm}" & M-C2  	& 41.20			& 40.02	& 40.61	\\ 
\hspace{0.5cm}" & M-C3  	& {\bf41.56}	& {\bf40.22}			& {\bf40.89}			\\ 
\hline
\end{tabular}
\caption{BLEU score on the test set of models trained with large-scale data, from English to Brazilian Portuguese (pt-BR) and European Portuguese (pt-EU).
Arrows $\downarrow\uparrow$ indicate statistically significant differences calculated against the \texttt{Mul} model.
}
\label{table:res_large}
\end{table}

In the semi-supervised scenario, the obtained results confirm that our approach of automatically classifying the unlabeled data $D_{E\rightarrow A \cup B}$ improves over using the data as they are ({\tt M-U}). Nevertheless, {\tt M-U} still confirms to perform better than the fully unlabeled {\tt Gen} model. In both translation directions, {\tt M-C2} and {\tt M-C3} get quite close to the performance of the supervised {\tt Spec} model. In particular, {\tt M-C3} shows to outperform the {\tt M-C2} model, and even outperforms on average the supervised {\tt Mul} model. In other words, the semi-supervised model leveraging three-class automatic labels (of $D_{E\rightarrow A \cup B}$) seems to perform better than the supervised model with two dialect labels. 
\replace{Looking at the results of the significance difference}{Besides the comparable BLEU scores}, the supervised ({\tt Spec} and {\tt Mul}) perform in statistically insignificant way against the best semi-supervised ({\tt M-C3}), although outperforming the unsupervised ({\tt Gen}) model.

This result raises the question if relabeling all the training data can be a better option than using a combination of manual and automatic labels. This issue is investigated in the next subsection.

\subsection*{Unsupervised Multilingual Models}
\noindent
As discussed in  Section 4.3, the language classifier for the large-data condition is trained on dialect-to-dialect parallel data that does not overlap with the NMT training data. This condition permits hence to investigate a fully unsupervised training condition. In particular, we assume that all the available training data is unlabeled and  create automatic language labels for all  47.2M sentences of pt-BR and 25.5M sentences of pt-EU (see Table~\ref{table:data}).  In a similar way as in\short{ large-scale} {Table~\ref{table:res_large}}, we keep the experimental setting of {\tt M-C2} and {\tt M-C3} models.

\begin{table}[t!]
\centering
\begin{tabular}{llrrr}
\hline
				&			& pt-BR			& pt-EU			& average		\\
\hline
Unsuper. 		& M-C2  	& 41.50			& {\bf 40.21} 	& 40.86 	\\ 
\hspace{0.5cm}" & M-C3  	& {\bf 41.66}	& 40.13			& {\bf 40.90}			\\ 
\hline
\end{tabular}
\caption{BLEU scores on the test set by large scale multi-lingual models trained under an unsupervised condition, where all the training data are labeled automatically.}
\label{table:res_large_fully_classified}
\end{table}

Table~\ref{table:res_large_fully_classified} reports the results of the multilingual models trained under the above described unsupervised condition. In comparison with the semi-supervised condition, both {\tt M-C2} and {\tt M-C3} show a slight performance improvement. In particular, the three-label {\tt M-C3} performs on average slightly better than the two-label {\tt M-C2} model. Actually, the little difference is justified by the fact that the classifier used the ``third'' label only for 6\% of the data. Remarkably, despite the relatively low performance of the classifier, average score of the best unsupervised model {\tt M-C2} is almost on par with the supervised model {\tt Mul}.

\subsection{Translation Examples}

Finally, in Table~\ref{table:examples_more}, we show an additional 
translation example produced by our semi-supervised multilingual  models (both under low and high resource conditions) translating into the Portuguese varieties. For comparison we also include output from Google Translate which offers only a generic English-Portuguese direction. In particular, the examples contain the word {\em refrigerator} that has specific dialect variants. All our variety-specific systems show to generate consistent translations of this term, while Google Translate prefers to use the Brazilian translation variants for these sentences. 

\begin{table*}[!t]
\centering
\small
\begin{tabular}{p {0.2 \textwidth} p{0.75 \textwidth}}
English (source)  & We offer a considerable number of different \underline{refrigerator} models. We have also developed a new type of \underline{ refrigerator}. These include American-style side-by-side \underline{refrigerators}. \\ [1mm] \hline \vspace{1mm}
pt (Google Translate) & \vspace{1mm} ferecemos um n\'umero consider\'avel de modelos diferentes de \textcolor{olive}{refrigeradores}. N\'os tamb\'em desenvolvemos um novo tipo de \textcolor{olive}{geladeira}. Estes incluem \textcolor{olive}{refrigeradores} lado a lado estilo americano. \\ [1mm]
{\bf Low-resource models} & \\ \hline \vspace{1mm}
pt-BR (M-C2) & \vspace{1mm} N\'os oferecemos um n\'umero consider\'avel de diferentes modelos de \textcolor{olive}{refrigerador}. Tamb\'em desenvolvemos um novo tipo de \textcolor{olive}{refrigerador}. Eles incluem o estilo americano nas \textcolor{olive}{geladeiras} lado a lado. \vspace{1mm} \\ \cdashline{2-2} \vspace{1mm}
pt-EU (M-C2) & \vspace{1mm} Oferecemos um n\'umero consider\'avel de modelos de \textcolor{blue}{refrigera\c{c}\~ao} diferentes. Tamb\'em desenvolvemos um novo tipo de \textcolor{blue}{frigor\'ifico}. Tamb\'em desenvolvemos um novo tipo de \textcolor{blue}{frigor\'ifico}. \\ [1mm] 
{\bf High-resource models} & \vspace{1mm} \\ \hline \vspace{1mm}
Spec-pt-BR & \vspace{1mm} Oferecemos um número consider\'avel de modelos de \textcolor{olive}{geladeira} diferentes. Tamb\'em desenvolvemos um novo tipo de \textcolor{olive}{geladeira}. Isso inclui o estilo americano lado a lado \textcolor{olive}{refrigeradores}. \vspace{1mm} \\  \cdashline{2-2} \vspace{1mm}
Spec-pt-PT & \vspace{1mm} Oferecemos um n\'umero consider\'avel de modelos de \textcolor{blue}{frigor\'ifico} diferentes. Tamb\'em desenvolvemos um novo tipo de \textcolor{blue}{frigorífico}. Estes incluem \textcolor{blue}{frigor\'ificos} americanos lado a lado. \vspace{2mm} \\ \cdashline{2-2} \vspace{1mm}
pt-BR (M-C3\_{\tt L}) & \vspace{1mm} Oferecemos um n\'umero consider\'avel de diferentes modelos de \textcolor{olive}{geladeira}. Tamb\'em desenvolvemos um novo tipo de \textcolor{olive}{geladeira}. Estes incluem estilo americano lado a lado, \textcolor{olive}{geladeiras}. \vspace{1mm} \\ \cdashline{2-2} \vspace{1mm} 
pt-PT (M-C3\_{\tt L}) &  \vspace{1mm} Oferecemos um n\'umero consider\'avel de diferentes modelos \textcolor{blue}{frigor\'ificos}. Tamb\'em desenvolvemos um novo tipo de \textcolor{blue}{frigorífico}. Estes incluem estilo americano lado a lado \textcolor{blue}{frigor\'ificos}. \\ \hline \vspace{1mm}
\end{tabular}
\caption{
English to Portuguese translation generated by Google Translate (as of 20th July, 2018) and translations into Brazilian and European Portuguese generated by \replace{one of our semi-supervised multilingual model ({\tt M-C2})}{our semi-supervised multilingual ({\tt M-C2} and {\tt M-C3\_L}) and supervised {\tt Spec} models}. For the underlined English terms both their \textcolor{olive}{Brazilian} and \textcolor{blue}{European} translation variants are shown.
}
\label{table:examples_more}
\end{table*}

\section{Conclusions}
We presented initial work on neural machine translation from English into dialects and related languages. We discussed both situations where parallel data is supplied or not supplied with target language/dialect labels. We introduced and compared different neural MT models that can be trained under unsupervised, supervised, and semi-supervised training data regimes. We reported  experimental results on the translation from English to four pairs of language varieties with systems trained under low-resource conditions. We show that in the supervised regime, best performance is achieved by training a multilingual NMT system. For the semi-supervised regime, we compared different automatic labeling strategies that permit to train multilingual neural MT systems with performance comparable to the best supervised NMT system. Our findings were also confirmed by large scale experiments performed on English to Brazilian and European Portuguese. In this scenario, we have also shown that multilingual NMT fully 
trained on automatic labels can perform very similarly to 
its supervised version. 

In future work, we plan to extend our approach to language varieties in the source side\short{ of an MT system}{}, as well as investigate the possibility of applying transfer-learning~\cite{zoph2016transfer,nguyen2017transfer} for language varieties by expanding our {\tt Ada} adaptation approach. 

\section*{Acknowledgments}
This work has been partially supported by the EC-funded project ModernMT (H2020 grant agreement no. 645487)\short{and QT21 (H2020 grant agreement no. 645452)}. We also gratefully acknowledge the support of NVIDIA Corporation with the donation of the Titan Xp GPU used for this research. Moreover, we thank the Erasmus Mundus European Program in Language and Communication Technology.

\bibliography{emnlp2018}

\begin{thebibliography}{37}
\expandafter\ifx\csname natexlab\endcsname\relax\def\natexlab#1{#1}\fi

\bibitem[{Altintas and {\c{C}}i{\c{c}}ekli(2002)}]{Altintas2002}
Kemal Altintas and İlyas {\c{C}}i{\c{c}}ekli. 2002.
\newblock {A Machine Translation System Between a Pair of Closely Related
  Languages}.
\newblock In \emph{Proceedings of the 17th International Symposium on Computer
  and Information Sciences (ISCIS 2002)}, pages 192--196.

\bibitem[{Bentivogli et~al.(2018)Bentivogli, Bisazza, Cettolo, and
  Federico}]{bentivogli:CSL2018}
Luisa Bentivogli, Arianna Bisazza, Mauro Cettolo, and Marcello Federico. 2018.
\newblock Neural versus phrase-based mt quality: An in-depth analysis on
  english-german and english-french.
\newblock \emph{Computer Speech \& Language}, 49:52--70.

\bibitem[{Cettolo et~al.(2012)Cettolo, Girardi, and
  Federico}]{cettoloEtAl:EAMT2012}
Mauro Cettolo, Christian Girardi, and Marcello Federico. 2012.
\newblock Wit$^3$: Web inventory of transcribed and translated talks.
\newblock In \emph{Proceedings of the 16$^{th}$ Conference of the European
  Association for Machine Translation (EAMT)}, pages 261--268, Trento, Italy.

\bibitem[{Costa-juss{\`{a}}(2017)}]{Costajussa2017}
Marta~R Costa-juss{\`{a}}. 2017.
\newblock {Why Catalan-Spanish Neural Machine Translation? Analysis, comparison
  and combination with standard Rule and Phrase-based technologies}.
\newblock In \emph{Proceedings of the Fourth Workshop on NLP for Similar
  Languages, Varieties and Dialects (VarDial)}, pages 55--62.

\bibitem[{Costa-Juss{\`{a}} et~al.(2018)Costa-Juss{\`{a}}, Zampieri, and
  Pal}]{Costa-Jussa2018}
Marta~R Costa-Juss{\`{a}}, Marcos Zampieri, and Santanu Pal. 2018.
\newblock {A Neural Approach to Language Variety Translation}.
\newblock In \emph{Proceedings of the Fifth Workshop on NLP for Similar
  Languages, Varieties and Dialects (VarDial)}, pages 275--282.

\bibitem[{Denkowski and Neubig(2017)}]{denkowski2017stronger}
Michael Denkowski and Graham Neubig. 2017.
\newblock Stronger baselines for trustable results in neural machine
  translation.
\newblock In \emph{Proceedings of the First Workshop on Neural Machine
  Translation}, pages 18--27.

\bibitem[{Dong et~al.(2015)Dong, Wu, He, Yu, and Wang}]{dong2015multi}
Daxiang Dong, Hua Wu, Wei He, Dianhai Yu, and Haifeng Wang. 2015.
\newblock Multi-task learning for multiple language translation.
\newblock In \emph{ACL (1)}, pages 1723--1732.

\bibitem[{Durrani et~al.(2010)Durrani, Sajjad, Fraser, and
  Schmid}]{durrani2010hindi}
Nadir Durrani, Hassan Sajjad, Alexander Fraser, and Helmut Schmid. 2010.
\newblock Hindi-to-urdu machine translation through transliteration.
\newblock In \emph{Proceedings of the 48th Annual meeting of the Association
  for Computational Linguistics}, pages 465--474. Association for Computational
  Linguistics.

\bibitem[{Fawcett(2006)}]{Fawcett2006}
Tom Fawcett. 2006.
\newblock {An introduction to ROC analysis}.
\newblock \emph{Pattern Recognition Letters}, 27(8):861--874.

\bibitem[{Firat et~al.(2016)Firat, Cho, and Bengio}]{firat2016multi}
Orhan Firat, Kyunghyun Cho, and Yoshua Bengio. 2016.
\newblock Multi-way, multilingual neural machine translation with a shared
  attention mechanism.
\newblock \emph{arXiv preprint arXiv:1601.01073}.

\bibitem[{Goutte et~al.(2016)Goutte, L{\'{e}}ger, Malmasi, and
  Zampieri}]{Goutte2016}
Cyril Goutte, Serge L{\'{e}}ger, Shervin Malmasi, and Marcos Zampieri. 2016.
\newblock {Discriminating Similar Languages: Evaluations and Explorations}.
\newblock In \emph{Proceedings of Language Resources and Evaluation (LREC)},
  pages 1800--1807.

\bibitem[{Ha et~al.(2016)Ha, Niehues, and Waibel}]{ha2016toward}
Thanh-Le Ha, Jan Niehues, and Alexander Waibel. 2016.
\newblock Toward multilingual neural machine translation with universal encoder
  and decoder.
\newblock \emph{arXiv preprint arXiv:1611.04798}.

\bibitem[{Harrat et~al.(2017)Harrat, Meftouh, and Smaili}]{Harrat2017}
Salima Harrat, Karima Meftouh, and Kamel Smaili. 2017.
\newblock {Machine translation for Arabic dialects (survey)}.
\newblock \emph{Information Processing \& Management}, pages 1--12.

\bibitem[{Hassan et~al.(2017)Hassan, Elaraby, and Tawfik}]{Hassan2017}
Hany Hassan, Mostafa Elaraby, and Ahmed~Y Tawfik. 2017.
\newblock {Synthetic Data for Neural Machine Translation of Spoken-Dialects}.
\newblock In \emph{Proceedings of the 14th International Workshop on Spoken
  Language Translation}.

\bibitem[{Jauhiainen et~al.(2018)Jauhiainen, Lui, Zampieri, Baldwin, and
  Lind{\'{e}}n}]{Jauhiainen2018}
Tommi Jauhiainen, Marco Lui, Marcos Zampieri, Timothy Baldwin, and Krister
  Lind{\'{e}}n. 2018.
\newblock {Automatic Language Identification in Texts: A Survey}.

\bibitem[{Johnson et~al.(2016)Johnson, Schuster, Le, Krikun, Wu, Chen, Thorat,
  Vi{\'e}gas, Wattenberg, Corrado et~al.}]{johnson2016google}
Melvin Johnson, Mike Schuster, Quoc~V Le, Maxim Krikun, Yonghui Wu, Zhifeng
  Chen, Nikhil Thorat, Fernanda Vi{\'e}gas, Martin Wattenberg, Greg Corrado,
  et~al. 2016.
\newblock Google's multilingual neural machine translation system: Enabling
  zero-shot translation.
\newblock \emph{arXiv preprint arXiv:1611.04558}.

\bibitem[{Joulin et~al.(2017)Joulin, Grave, Bojanowski, and
  Mikolov}]{Joulin2017}
Armand Joulin, Edouard Grave, Piotr Bojanowski, and Tomas Mikolov. 2017.
\newblock Bag of tricks for efficient text classification.
\newblock In \emph{Proceedings of the 15th Conference of the European Chapter
  of the Association for Computational Linguistics: Volume 2, Short Papers},
  volume~2, pages 427--431.

\bibitem[{Kingma and Ba(2014)}]{kingma2014adam}
Diederik Kingma and Jimmy Ba. 2014.
\newblock Adam: A method for stochastic optimization.
\newblock \emph{arXiv preprint arXiv:1412.6980}.

\bibitem[{Koehn(2004)}]{Koehn2004}
Philipp Koehn. 2004.
\newblock {Statistical significance tests for machine translation evaluation}.
\newblock In \emph{Proceedings of the Conference on Empirical Methods in
  Natural Language Processing}, volume~4, pages 388--395.

\bibitem[{Koehn and Knowles(2017)}]{koehn2017six}
Philipp Koehn and Rebecca Knowles. 2017.
\newblock Six challenges for neural machine translation.
\newblock \emph{arXiv preprint arXiv:1706.03872}.

\bibitem[{Luong et~al.(2015)Luong, Le, Sutskever, Vinyals, and
  Kaiser}]{luong2015multi}
Minh-Thang Luong, Quoc~V Le, Ilya Sutskever, Oriol Vinyals, and Lukasz Kaiser.
  2015.
\newblock Multi-task sequence to sequence learning.
\newblock \emph{arXiv preprint arXiv:1511.06114}.

\bibitem[{Malmasi et~al.(2016)Malmasi, Zampieri, Ljube{\v{s}}i, Nakov, Ali, and
  Tiedemann}]{Malmasi2016}
Shervin Malmasi, Marcos Zampieri, Nikola Ljube{\v{s}}i, Preslav Nakov, Ahmed
  Ali, and J{\"{o}}rg Tiedemann. 2016.
\newblock {Discriminating Between Similar Languages and Arabic Dialect
  Identification: A Report on the Third DSL Shared Task}.
\newblock In \emph{Proceedings of the Third Workshop on NLP for Similar
  Languages, Varieties and Dialects}, pages 1--14.

\bibitem[{Marujo et~al.(2011)Marujo, Grazina, Luis, Ling, Coheur, and
  Trancoso}]{Marujo2011}
Luis Marujo, Nuno Grazina, Tiago Luis, Wang Ling, Luisa Coheur, and Isabel
  Trancoso. 2011.
\newblock {{BP2EP} - Adaptation of {B}razilian {P}ortuguese texts to {E}uropean
  {P}ortuguese}.
\newblock In \emph{Proceedings of the 15th International Conference of the
  European Association for Machine Translation (EAMT)}, May, pages 129--136.

\bibitem[{Nguyen and Chiang(2017)}]{nguyen2017transfer}
Toan~Q Nguyen and David Chiang. 2017.
\newblock Transfer learning across low-resource, related languages for neural
  machine translation.
\newblock In \emph{Proceedings of the Eighth International Joint Conference on
  Natural Language Processing (Volume 2: Short Papers)}, volume~2, pages
  296--301.

\bibitem[{Popovi{\'{c}} et~al.(2016)Popovi{\'{c}}, Arcan, and
  Klubi{\v{c}}ka}]{Popovic2016}
Maja Popovi{\'{c}}, Mihael Arcan, and Filip Klubi{\v{c}}ka. 2016.
\newblock {Language Related Issues for Machine Translation between Closely
  Related South Slavic Languages}.
\newblock In \emph{Proceedings of the Third Workshop on NLP for Similar
  Languages, Varieties and Dialects (VarDial3)}, pages 43--52.

\bibitem[{Pourdamghani and Knight(2017)}]{Pourdamghani2017}
Nima Pourdamghani and Kevin Knight. 2017.
\newblock {Deciphering Related Languages}.
\newblock In \emph{Proceedings of the 2017 Conference on Empirical Methods in
  Natural Language Processing}, pages 2503--2508.

\bibitem[{Salloum et~al.(2014)Salloum, Elfardy, Alamir-Salloum, Habash, and
  Diab}]{Salloum2014}
Wael Salloum, Heba Elfardy, Linda Alamir-Salloum, Nizar Habash, and Mona Diab.
  2014.
\newblock {Sentence Level Dialect Identification for Machine Translation System
  Selection}.
\newblock In \emph{Proceedings of the 52nd Annual Meeting of the Association
  for Computational Linguistics (Short Papers)}, pages 772--778.

\bibitem[{Tan et~al.(2012)Tan, Goh, and Khaw}]{Tan2012}
Tien-Ping Tan, Sang-Seong Goh, and Yen-Min Khaw. 2012.
\newblock {A Malay Dialect Translation and Synthesis System: Proposal and
  Preliminary System}.
\newblock In \emph{2012 International Conference on Asian Language Processing},
  pages 109--112. IEEE.

\bibitem[{Tiedemann and Ljube{\v{s}}i(2012)}]{Tiedemann2012}
J{\"{o}}rg Tiedemann and Nikola Ljube{\v{s}}i. 2012.
\newblock {Efficient Discrimination Between Closely Related Languages}.
\newblock In \emph{Proceedings of COLING 2012: Technical Papers}, pages
  2619--2634.

\bibitem[{Vaswani et~al.(2018)Vaswani, Bengio, Brevdo, Chollet, Gomez, Gouws,
  Jones, Kaiser, Kalchbrenner, Parmar, Sepassi, Shazeer, and
  Uszkoreit}]{tensor2tensor}
Ashish Vaswani, Samy Bengio, Eugene Brevdo, Francois Chollet, Aidan~N. Gomez,
  Stephan Gouws, Llion Jones, \L{}ukasz Kaiser, Nal Kalchbrenner, Niki Parmar,
  Ryan Sepassi, Noam Shazeer, and Jakob Uszkoreit. 2018.
\newblock Tensor2tensor for neural machine translation.
\newblock \emph{CoRR}, abs/1803.07416.

\bibitem[{Vaswani et~al.(2017)Vaswani, Shazeer, Parmar, Uszkoreit, Jones,
  Gomez, Kaiser, and Polosukhin}]{vaswani2017attention}
Ashish Vaswani, Noam Shazeer, Niki Parmar, Jakob Uszkoreit, Llion Jones,
  Aidan~N Gomez, {\L}ukasz Kaiser, and Illia Polosukhin. 2017.
\newblock Attention is all you need.
\newblock In \emph{Advances in Neural Information Processing Systems}, pages
  6000--6010.

\bibitem[{Wardhaugh(2006)}]{Wardhaugh}
Ronald Wardhaugh. 2006.
\newblock \emph{An Introduction to Sociolinguistcs}.
\newblock Blackwell Publishing.

\bibitem[{Wu et~al.(2016)Wu, Schuster, Chen, Le, Norouzi, Macherey, Krikun,
  Cao, Gao, Macherey et~al.}]{wu2016google}
Yonghui Wu, Mike Schuster, Zhifeng Chen, Quoc~V Le, Mohammad Norouzi, Wolfgang
  Macherey, Maxim Krikun, Yuan Cao, Qin Gao, Klaus Macherey, et~al. 2016.
\newblock Google's neural machine translation system: Bridging the gap between
  human and machine translation.
\newblock \emph{arXiv preprint arXiv:1609.08144}.

\bibitem[{Zampieri et~al.(2017)Zampieri, Malmasi, Ljube{\v{s}}i, Nakov, Ali,
  Tiedemann, Scherrer, and Aepli}]{Zampieri2017}
Marcos Zampieri, Shervin Malmasi, Nikola Ljube{\v{s}}i, Preslav Nakov, Ahmed
  Ali, J{\"{o}}rg Tiedemann, Yves Scherrer, and No{\"{e}}mi Aepli. 2017.
\newblock {Findings of the VarDial Evaluation Campaign 2017}.
\newblock In \emph{Proceedings of the Fourth Workshop on NLP for Similar
  Languages, Varieties and Dialects}, pages 1--15.

\bibitem[{Zampieri et~al.(2014)Zampieri, Tan, Ljube{\v{s}}i{\'{c}}, Tiedemann,
  and Ljubeˇ}]{Zampieri2014}
Marcos Zampieri, Liling Tan, Nikola Ljube{\v{s}}i{\'{c}}, J{\"{o}}rg Tiedemann,
  and Nikola Ljubeˇ. 2014.
\newblock {A Report on the DSL Shared Task 2014}.
\newblock In \emph{Proceedings of the First Workshop on Applying NLP Tools to
  Similar Languages, Varieties and Dialects}, 2013, pages 58--67.

\bibitem[{Zampieri et~al.(2015)Zampieri, Tan, Ljube{\v{s}}i{\'{c}}, Tiedemann,
  and Nakov}]{Zampieri2015}
Marcos Zampieri, Liling Tan, Nikola Ljube{\v{s}}i{\'{c}}, J{\"{o}}rg Tiedemann,
  and Preslav Nakov. 2015.
\newblock {Overview of the DSL Shared Task 2015}.
\newblock In \emph{Proceedings of the Joint Workshop on Language Technology for
  Closely Related Languages, Varieties and Dialects}, 2014, pages 1--9.

\bibitem[{Zoph et~al.(2016)Zoph, Yuret, May, and Knight}]{zoph2016transfer}
Barret Zoph, Deniz Yuret, Jonathan May, and Kevin Knight. 2016.
\newblock Transfer learning for low-resource neural machine translation.
\newblock \emph{arXiv preprint arXiv:1604.02201}.

\end{thebibliography}
\bibliographystyle{acl_natbib_nourl}

\end{document}